\documentclass{article}
\usepackage[utf8]{inputenc}
\usepackage{authblk}
\usepackage{setspace}
\usepackage[margin=1.25in]{geometry}
\usepackage{graphicx}
\graphicspath{ {./figures/} }
\usepackage{subcaption}
\usepackage{amsmath}
\usepackage{subcaption}
\usepackage{multirow}
\usepackage{graphicx}
\usepackage{array}
\usepackage{hyperref}
\usepackage[style=nejm, 
citestyle=numeric-comp,
sorting=none]{biblatex}
\title{Trees as Gaussians: \\Large-Scale Individual Tree Mapping}

\author[1]{Dimitri Gominski}
\author[1]{Martin Brandt}
\author[1]{Xiaoye Tong}
\author[1]{Siyu Liu}
\author[1]{Maurice Mugabowindekwe}
\author[1]{Sizhuo Li}
\author[1]{Florian Reiner}
\author[2]{Andrew Davies}
\author[1]{Rasmus Fensholt}

\affil[1]{University of Copenhagen}
\affil[2]{Harvard University}
\date{}

\begin{document}

\maketitle

%%%%%% Abstract %%%%%%
\begin{abstract}
Trees are key components of the terrestrial biosphere, playing vital roles in ecosystem function, climate regulation, and the bioeconomy. However, large-scale monitoring of individual trees remains limited by inadequate modelling. Available global products have focused on binary tree cover or canopy height, which do not explicitely identify trees at individual level. In this study, we present a deep learning approach for detecting large individual trees in 3-m resolution PlanetScope imagery at a global scale. We simulate tree crowns with Gaussian kernels of scalable size, allowing the extraction of crown centers and the generation of binary tree cover maps. Training is based on billions of points automatically extracted from airborne lidar data, enabling the model to successfully identify trees both inside and outside forests. We compare against existing tree cover maps and airborne lidar with state-of-the-art performance (fractional cover R$^2 = 0.81$ against aerial lidar), report balanced detection metrics across biomes, and demonstrate how detection can be further improved through fine-tuning with manual labels. Our method offers a scalable framework for global, high-resolution tree monitoring, and is adaptable to future satellite missions offering improved imagery.
\end{abstract}

%%%%%% Main Text %%%%%%

\section{Introduction}
Trees represent an essential component of the terrestrial biosphere. They cover large parts of the land surface and are essential for the functioning of most ecosystems by contributing to ecosystem services such as climate regulation, fertilization, soil stabilization and biodiversity support \cite{wang_biodiversity_2019}. Moreover, they provide economic benefits for local livelihoods, in the form of nutrition, timber, fuel, construction wood and other wood materials for the bioeconomy \cite{rosenstock_planetary_2019}. %Changes in the number, cover and composition of trees have a fundamental impact on the bio-geosphere. 
Both human management and climate change have profoundly affected tree cover and composition over the past centuries. The exploitation of forest resources generates economic income, but releases CO$_2$ and transforms habitats, highlighting the importance of monitoring to prepare the transition to more sustainable practices \cite{mugabowindekwe_trees_2024}.

The monitoring of trees at large scale is typically done with satellite data and models have been developed to estimate binary cover (tree/no tree) \cite{hansen_high-resolution_2013, brandt_wall--wall_2023} or canopy height \cite{lang_high-resolution_2023, tolan_very_2024, pauls_estimating_2024}. Images from Landsat and Sentinel 1 and 2 satellites have been used to map tree cover, however, the spatial resolution of $\geq10$ m restricts the mapping to closed canopy areas or groups of trees, thereby omitting many trees outside forests. Moreover, it remains challenging to monitor subtle dynamics in tree cover, restricting focus on dynamics that can be reliably mapped such as clear cuts at the stand level or larger forest areas \cite{dalagnol_mapping_2023}. With the advent of deep learning deployed on very high resolution satellite imagery, mapping at the level of single trees is feasible \cite{tolan_very_2024}. Sub-meter images from commercial sources have sucesfully been used under the nextview licence agreement to segment individual tree crowns at the subcontinental scale, with large volumes of hand-labeled polygons for training dedicated models \cite{brandt_unexpectedly_2020}. From licensed PlanetScope data access images at 3-m resolution, it is also possible to segment trees that have reached a minimum size, estimated to be between 10 and 30 m$^2$ \cite{reiner_more_2023}.

However, global canopy height and cover products do not explicitly identify single trees. Identifying trees as single objects is a paradigm shift, as it makes their number countable, enables locating individual trees both inside and outside forests, and allows tracking over time \cite{mugabowindekwe_nation-wide_2023, brandt_severe_2024}. %In the absense of sub-meter resolution data, separating trees into individuals has been challenging, in particular for canopy areas of dense tree cover with overlapping tree crowns. In such cases manual labelling is not possible, as the spatial resolution is not high enough to clearly identify the boundaries of single trees, being the case e.g. for PlanetScope imagery. Manual labels and canopy height maps from airborne lidar have been used to train deep learning models that produce binary tree/no-tree segmentation maps from PlanetScope images \cite{reiner_more_2023, liu_overlooked_2023}. However, these maps often miss trees outside forests, and tend to group trees into clusters of trees and consequently it is hardly possible to derive information on the location of single trees, in particular not in forests where trees are mapped as continuous cover. %Our approach builds on recent advances in vision models with deep learning. Convolutional neural networks have shown outstanding performance for learning functions mapping input data (here, satellite imagery) to a desired output. The typical setup involves a feature extractor composed of a series of convolutional blocks and activation functions, and one or multiple heads converting the resulting “deep features” to outputs. While there are well-established generalistic feature extractors such as UNet or ResNet, output head design is specific to the task at hand. 
Object detection is a well-studied subject in computer vision. Popular approaches can be separated into two broad families. \textit{Anchor-based detection} maps the input to a list of outputs associated with predefined locations; the anchors. Each anchor is classified to indicate presence/absence of objects at that location. This has been applied successfully for single tree detection on high resolution imagery \cite{weinstein_remote_2021, g_braga_tree_2020}. A major downside is the fixed grid of anchors, which largely influences the minimum mappable object size and maximum object number/density. It is also an issue for running large-scale predictions, as inference is typically done at larger image sizes than training to speed up prediction speed. With a fixed grid, it is not possible to change the image size fed to the model. \textit{Anchor-free detection} removes the dependence on the anchor grid, typically by transforming the problem into a regression task where each pixel is assigned a probability of containing an object keypoint \cite{zhou_objects_2019, law_cornernet_2020}. These pseudo-probability maps, or heatmaps, have the advantage of making outputs easy to interpret and postprocess, and if using a fully convolutional network, can be predicted for any image size. Anchor-free tree detection has been applied with success at national-scale in India \cite{brandt_severe_2024}.

\begin{figure}[h]
    \centering
    \includegraphics[width=\linewidth]{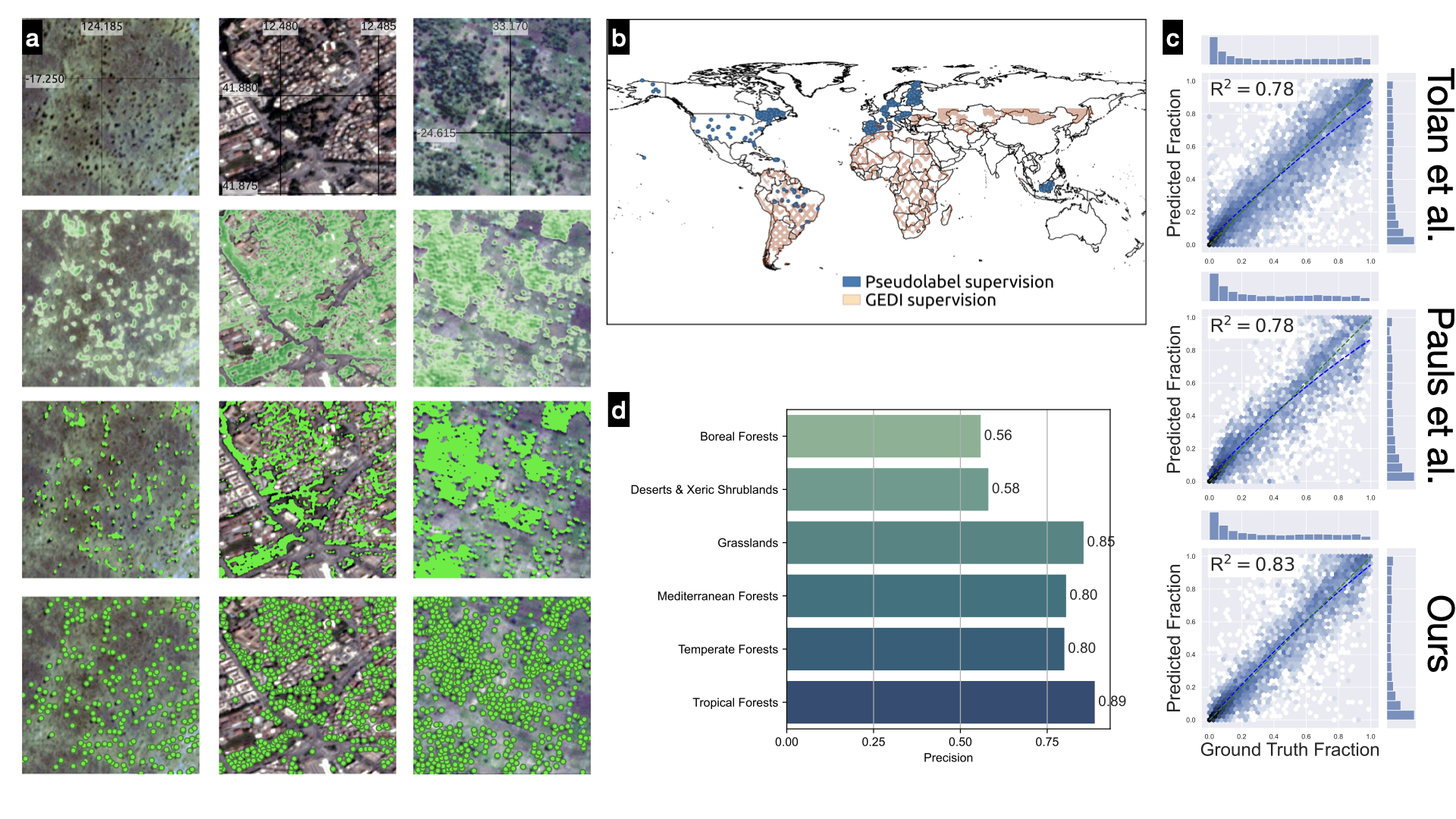}
    \caption{A generalistic model for mapping and detecting trees down to individual level, based on anchor-free detection with heatmaps \textbf{(a)}. Heatmaps (second row) can be thresholded, giving binary cover maps (third row), or processed into tree positions with local maxima detection (fourth row). With a curated but mostly automated mining of labels, we build a global training dataset \textbf{(b)}, leading to competitive performance for tree cover mapping \textbf{(c)} and detection metrics highlighting a strong capacity to generalize to various biomes \textbf{(d)}.}
    \label{fig:mainfig}
\end{figure}

This study presents a global training approach leveraging models that are able to identify trees in PlanetScope images, both inside and outside forests. Here, we present an anchor-free approach that we believe constitutes a good compromise between scale, accuracy and cost-effectiveness. We simulate tree crowns with gaussians that scale with the crown size. The peak value of the gaussian can be used to extract the tree crown center locations, and a threshold can be chosen to convert the map into a binary tree cover map (see Figure~\ref{fig:mainfig}). Models are trained with billions of points automatically extracted from lidar data globally, enabling the identification of trees that are hardly visible for the human eye. The spatial resolution of 3-m restricts the mapping to larger trees. In dense forests where trees have small crowns (e.g. coniferous forests), trees are grouped into clusters, which is a clear limitation of the data source. The approach is, however, transfereable to future satellite missions that may provide global coverage of higher quality images as compared to PlanetScope in regard to spatial and radiometric resolution. We demonstrate how our model performs in selected areas globally for tree detection and cover mapping, and we compare our maps with existing state-of-the-art tree cover maps. Finally, we show how our predictions can be fine-tuned with manual labels to increase performance.  

\section{Materials and Methods}
\subsection{Gaussian Modeling}
\label{subsec:gaussian_modeling}

We represent trees as Gaussians: tree centers generate a Gaussian kernel at position (x, y) with standard deviation $\sigma$ and amplitude 1. The kernels are max-pooled to give a target output at full resolution, the heatmap, which can be seen as a probability density function indicating for each pixel the probability that it contains a tree center. Our model has two heads, the heatmap head and a head modelling spatial uncertainty. Gaussian kernels are resized during training using the output of the spatial uncertainty head: if there is high uncertainty about the position of tree centers in a certain area, Gaussian kernels are enlarged, which increases the chances that kernels overlap with tree crowns and helps convergence. On the contrary, if there is low uncertainty about center position, kernels are drawn with a minimal, default $\sigma$ value. In practice, Gaussian kernels are drawn online during training, using labeled tree positions, $\sigma$, and the predicted spatial uncertainty. Initially, all kernels are drawn with the same size, as the spatial uncertainty head is initialized to output zero. Then during optimization, the model learns to use spatial uncertainty to modulate kernel size locally. We train our model for pixel regression using the following loss function: 

\begin{equation}
    L_{total} = L_{MSE}(h, y) + \delta ||s||^2,
\end{equation}

where the first term encourages prediction of the drawn heatmaps and the second term acts as a regularizer preventing the collapse to high spatial uncertainty everywhere. $h$ indicates the predicted heatmap, $y$ the online-drawn Gaussian target, $s$ the predicted spatial uncertainty.

Our model trains with point labels. It combines the accuracy of pixel-based approaches with enhanced scalability, point labels being much easier to collect than individual polygons. Uncertainty modelling allows training with different crown sizes seamlessly, and adds robustness to label noise.

\subsection{Generating Pseudo-Labels}
\label{subsec:generating_labels}

We introduce a cost-effective way to collect large volumes of point pseudo-labels from Canopy Height Models derived from aerial lidar campaigns, with limited human involvement and high scalability. 

\begin{figure}[h]
    \centering
    \includegraphics[width=0.7\linewidth]{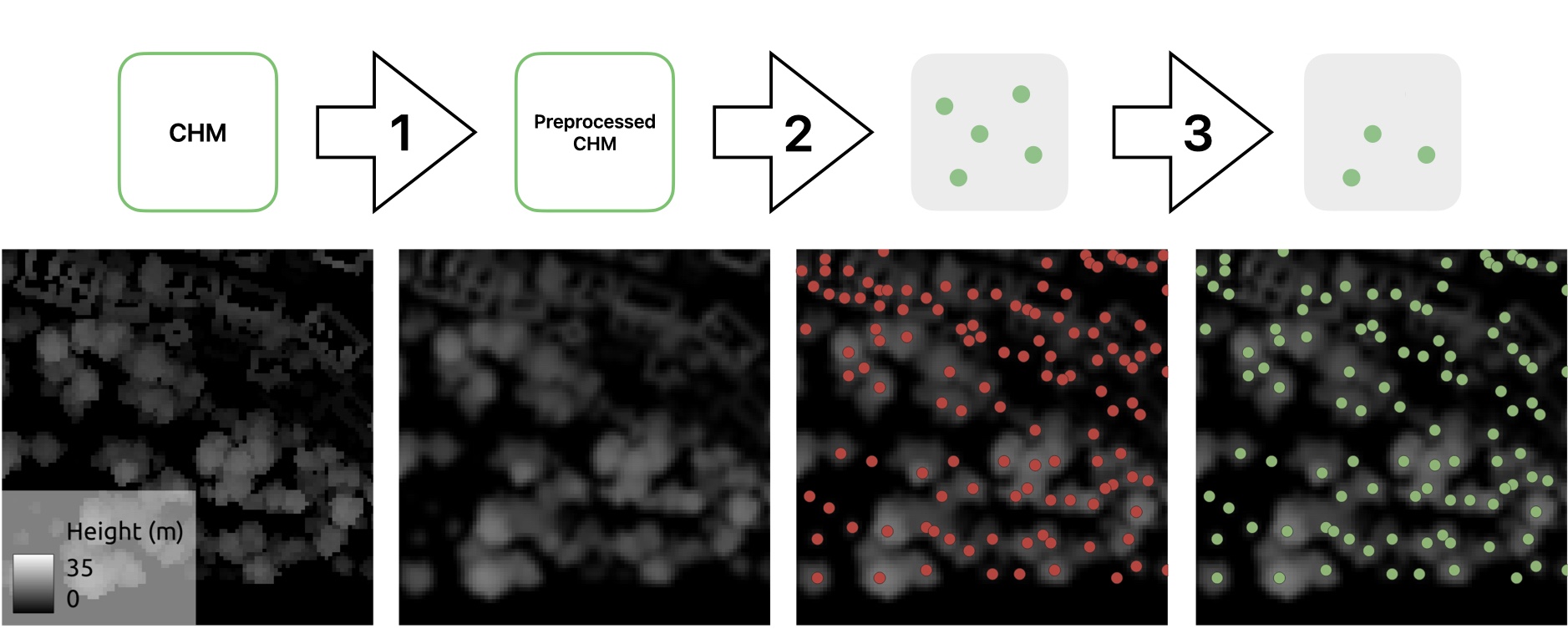}
    \caption{Point extraction pipeline. 1- Canopy Height Models (CHMs) are preprocessed to limit quality issues and maximize detection performance. 2- Tree centers are detected as peak locations on the preprocessed CHM. 3- Detections are post-processed to remove points on buildings or with low height values. The second row displays an example with CHMs in Slovenia: preprocessing smoothes artifacts created by wrongly classified lidar points, and postprocessing removes points that fall within a building footprint in the top portion.}
    \label{fig:chm_pipeline}
\end{figure}

We extract tree centers as the positions of local maxima on CHMs. Our pipeline, represented in Figure~\ref{fig:chm_pipeline}), involves three important steps (\textit{and associated parameters}). First, CHMs are preprocessed with a set of simple image operations : average blurring (\textit{kernel size}), median blurring (\textit{kernel size}), Gaussian blurring (\textit{kernel size, sigma}), and a parameterized version of the CHM correction approach from Deng et al. \cite{deng_comparison_2016} (\textit{kernel size, number of filtering steps, high threshold, low threshold, blurring kernel size, blurring sigma}). Once preprocessed, we perform detection on the CHMs, by identifying peak values above a threshold on a sliding window that has either a fixed size (\textit{window size,  threshold}), or a size that varies depending on local height (\textit{minimum window size, maximum window size, height-to-window size factor, threshold}). 

Our approach involves 12 to 14 parameters. To limit bias and maximize the quality of the pseudo-labels, we automatically select the best parameters with Bayesian optimization. For each CHM collection, we manually label a small set of tree centers ($<5000$, taking up to 2hrs for a trained annotator), and run optimization with the Ax API to maximize the harmonic mean of the F1 score (5m maximum distance) and counting accuracy (computed as $1 - (|N-M|/N)$, with $N$ the number of detections and $M$ the number of labels). This ensures a balance between estimating a correct overall number of trees and predicting near labeled centers. 

The labels are also used to estimate the distribution of intercenter distances, which we plot against height values to choose the most suited approach. If there is a clear correlation between intercenter distance and height, we choose height-adaptive detection, otherwise fixed-window.

Our pseudo-label dataset contains approx. 14 billion trees and covers $1030000\text{ km}^2$, split into $950000 \text{ km}^2$ for training and $80000\text{ km}^2$ for validation .

% 1025139km2= 948536 + 76603 (12,253,825,326 training trees) (1,494,521,856 testing trees)

% \begin{figure}
%     \centering
%     \includegraphics[width=\linewidth]{ims/trainset_gedi.png}
%     \caption{Distribution of our train set of pseudo-labels extracted from CHMs and negative GEDI supervision.}
%     \label{fig:trainset}
% \end{figure}

\subsection{Architecture}

We train a UNet model \cite{ronneberger_u-net_2015} with a ResNet50 encoder \cite{he_deep_2016}, and two output heads to regress the heatmaps and positional uncertainty. Heads are a simple stack of two convolutions with activation to reduce the full resolution, 16-dimensional UNet features to a single channel output.

\subsection{Enhancing Generalization}

Despite efforts to collect data everywhere possible, our training dataset remains spatially biased. To alleviate performance imbalances and enhance generalization, we add two important components: learning geographic priors from location embeddings, and negative GEDI supervision.

\subsubsection{Geographic Priors}

Recent advances in Earth Observation foundation models have established new ways of extracting high-level knowledge for various applications, including tree mapping. They are typically trained on large amounts of data, with the goal of producing generalistic representations of places on Earth. This knowledge can be extracted by learning pretext tasks on various sources of image data with self-supervised learning \cite{astruc_omnisat_2025}, or by learning from semantic clues with supervised learning \cite{yin_gps2vec_2019}.

We use a method that belongs to the second category, SatCLIP \cite{klemmer_satclip_2024}. The model was trained with the CLIP paradigm \cite{radford_learning_2021} to align location encodings with a global Sentinel-2 dataset, making the model learn compact representations of places. For a given (longitude, latitude) input, SatCLIP produces an embedding. We inject the 256-dimensional embeddings from the ResNet50-L40 version in our model with a convolution layer merging them with the deep, low-resolution UNet features before the decoder. For a given patch, we compute one SatCLIP embedding corresponding to the patch center's longitude and latitude and expand it to the whole patch before the learnable convolution.

\subsubsection{Negative GEDI Supervision}

The Global Ecosystem Dynamics Investigation (GEDI) mission produces global (between 52\textdegree N and 52\textdegree S) estimates of canopy height from satellite lidar technology. Its consistency and homegeneity makes it a fitting source of training data for canopy height models \cite{lang_high-resolution_2023, pauls_estimating_2024}. GEDI samples are delivered as waveforms captured on a circular sampling footprint, and processed to estimate forest structure. 

The coarse resolution of $~30$m is too low to provide positive supervision for individual tree mapping. However, on locations where the waveform does not indicate clear peaks (peaks appear when pulses bounce on vegetation or other structures), we can assume that there is no tree present and use it as a form of negative supervision.

We sample GEDI waveforms from the 2A product, focusing on areas where our training data is limited, notably Africa and South America. We identify negative footprints when the following conditions are met: the number of detected modes is equal to one (the ground), sensitivity is above 0.9 (indicating a reliable sample), and the estimated RH98 is below 2.5m. We plot the spatial distribution of our negative GEDI supervision set in Figure~\ref{fig:mainfig}.

\subsection{Performance Evaluation}
\label{subsec:performance_eval}

We evaluate the capacities of large-scale models trained with Gaussian modeling through two downstream tasks, individual tree detection and cover mapping.

\subsubsection{Tree Detection}
\label{subsubsec:tree_detection}

\begin{figure}
\begin{subfigure}[T]{0.5\textwidth}
    \centering
    \includegraphics[width=\textwidth]{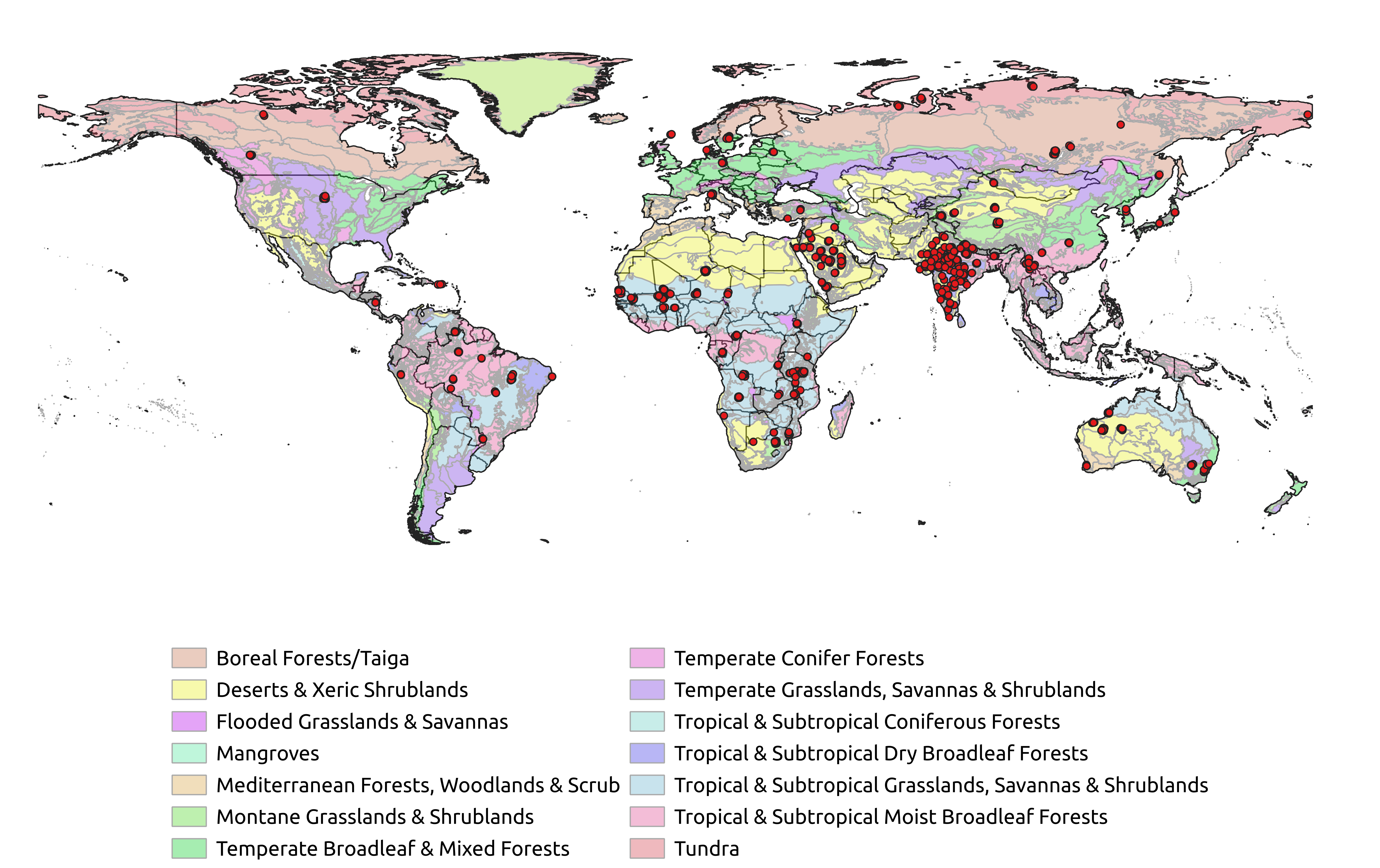}
    \caption{Tree detection, with manual labels made on PlanetScope imagery.}
        \label{fig:dettestset}

\end{subfigure}
\begin{subfigure}[T]{0.47\textwidth}
    \includegraphics[width=\textwidth]{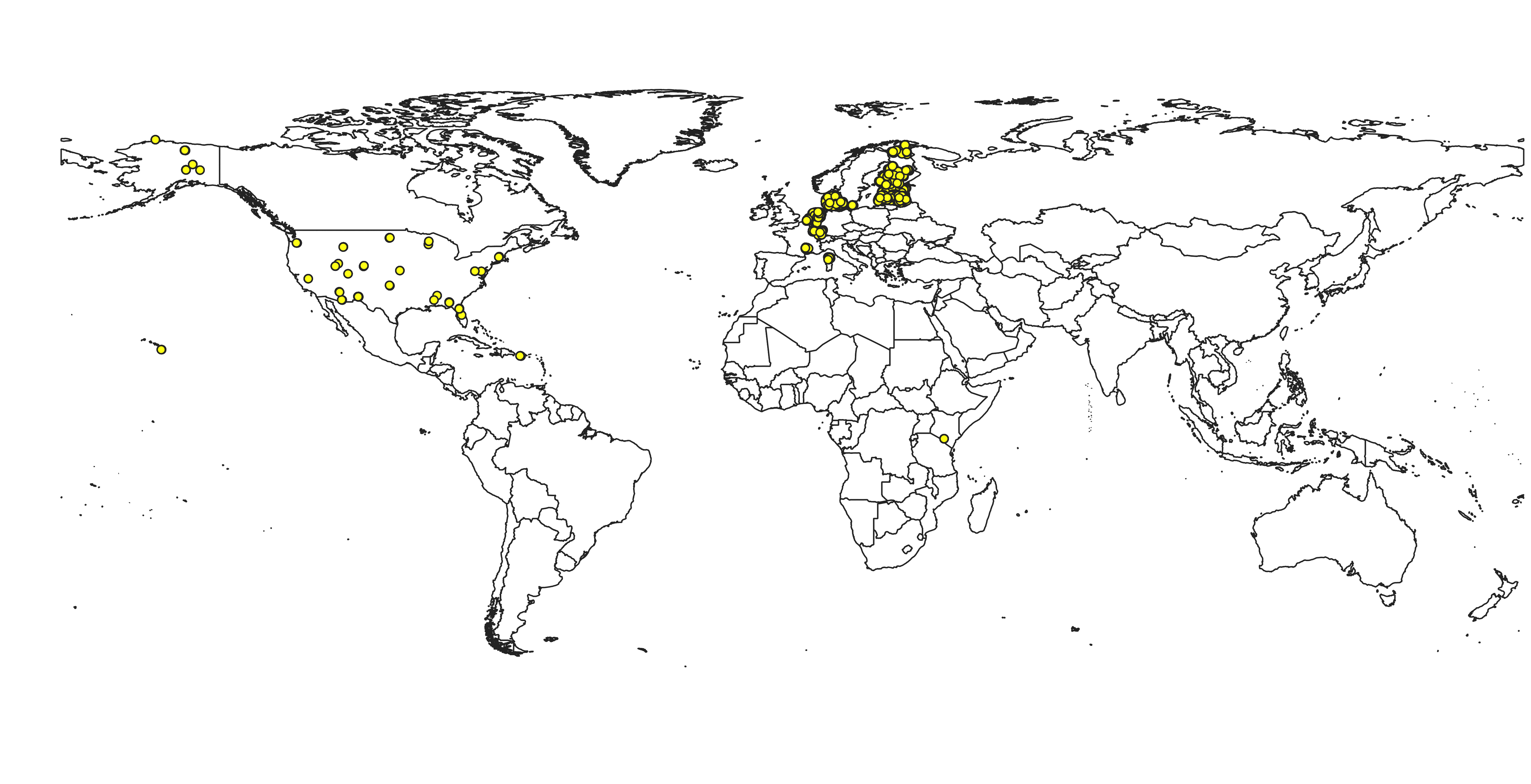}
    \caption{Cover mapping with canopy height models derived from aerial lidar (submeter resolution).}
    \label{fig:covertestset}
\end{subfigure}
\caption{Spatial distribution of our evaluation data.}
    \label{fig:testsets}

\end{figure}

We assess the capacity of models to detect trees that are visible in PlanetScope imagery. We collect a dataset of hand-drawn point labels made through photointerpretation of the satellite images. The dataset is based on Brandt et al., (2024) and was expanded to approx. 2 million trees on all continents. The labels were made with the intent of building a training set for individual tree detection, as such they focus on areas where individual trees can be unambiguously labeled by well-trained geographers, with various degrees of image quality. We stratify labeled areas by biomes using the map from (Dinerstein et al., 2017).

% \subsubsection{Detection of Smaller Trees}

% \begin{figure}
%     \centering
%     \includegraphics[width=\linewidth]{ims/globalbench.png}
%     \caption{Distribution of our test set for general tree detection, with manual labels made on sub-meter imagery.}
%     \label{fig:enter-label}
% \end{figure}

% We assess the capacity of models to detect smaller trees, whether or not they appear clearly on PlanetScope imagery. We collect a dataset of hand-drawn point labels made through photointerpretation of sub-meter WorldView imagery provided by Google Maps. The dataset contains 50.6 thousand trees drawn on randomly selected images for each continent. Here, we emphasize independent assessment by labeling trees of all sizes and densities, and disregard potential misalignment or quality issues with the satellite imagery.

\subsubsection{Cover Mapping}

We assess the capacity of models to map tree cover, defined as any pixel containing vegetation height above 3m. We collect a dataset of Canopy Height Models (CHMs) derived from aerial lidar campaigns, by keeping aside a portion of the CHMs used for our training dataset. This set was randomly selected with a stratified sampling of the countries for which we have such data either at national scale or on targeted areas. For each country, we randomly sample 2\% of the frames to create an initial test set. We add any frame that intersects a test frame to the test set, then crop test frames to the train set spatial extent to ensure zero overlap. This dataset is mechanically biased towards countries where lidar campaigns are led at national scale, mainly European and North American countries. 

To alleviate image quality issues and temporal mismatches between evaluation CHMs and PlanetScope images, we compute predictions of our best performing model for a stack of 2 to 5 images, and aggregate them with pixel-wise median.

We compute pixel-level (at PlanetScope resolution) metrics by resampling CHMs with nearest neighbor interpolation, and comparing with our predictions thresholded with different values.

We compare our predictions against five popular tree mapping products:  European Space Agency's WorldCover  (\cite{esa_worldcover_2020}, year 2020, based on Sentinel-1 and 2 imagery, 10m GSD), Joint Research Center's (JRC) Global Forest Cover (\cite{bourgoin_mapping_2024},  2020, composite product, 10m GSD), Lang et al. \cite{lang_high-resolution_2023} (2020, Sentinel-2, 10m GSD), Pauls et al. \cite{pauls_estimating_2024} (2019-2020, Sentinel-1 and 2, 10m GSD), Tolan et al. \cite{tolan_very_2024} (2018-2020, Maxar, 1m GSD). For a fair evaluation, we follow the same protocol for all products, including our model. For all sites where both product and CHM data are available, we collect it, filter out data windows smaller than 25mx25m or containing mostly nodata pixels, then run a parameter sweep to identify the threshold giving the best R$^2$ score between product cover fraction and CHM cover fraction (height threshold for height products, confidence threshold for ours, no threshold for WorldCover). We compute results with universal thresholds, as well as country-specific thresholds.

\section{Results and Discussion}

In the following, we evaluate our models trained through different setups. \textit{base} is the base model, initialized with ImageNet weights and trained on the set of pseudolabels (Figure~\ref{fig:mainfig}). \textit{+gedi} indicates the addition of negative GEDI supervision in the training set, while \textit{+satclip} indicates the use of SatCLIP embeddings.

\subsubsection{Tree Detection}

\begin{figure}
    \centering
    \includegraphics[width=\linewidth]{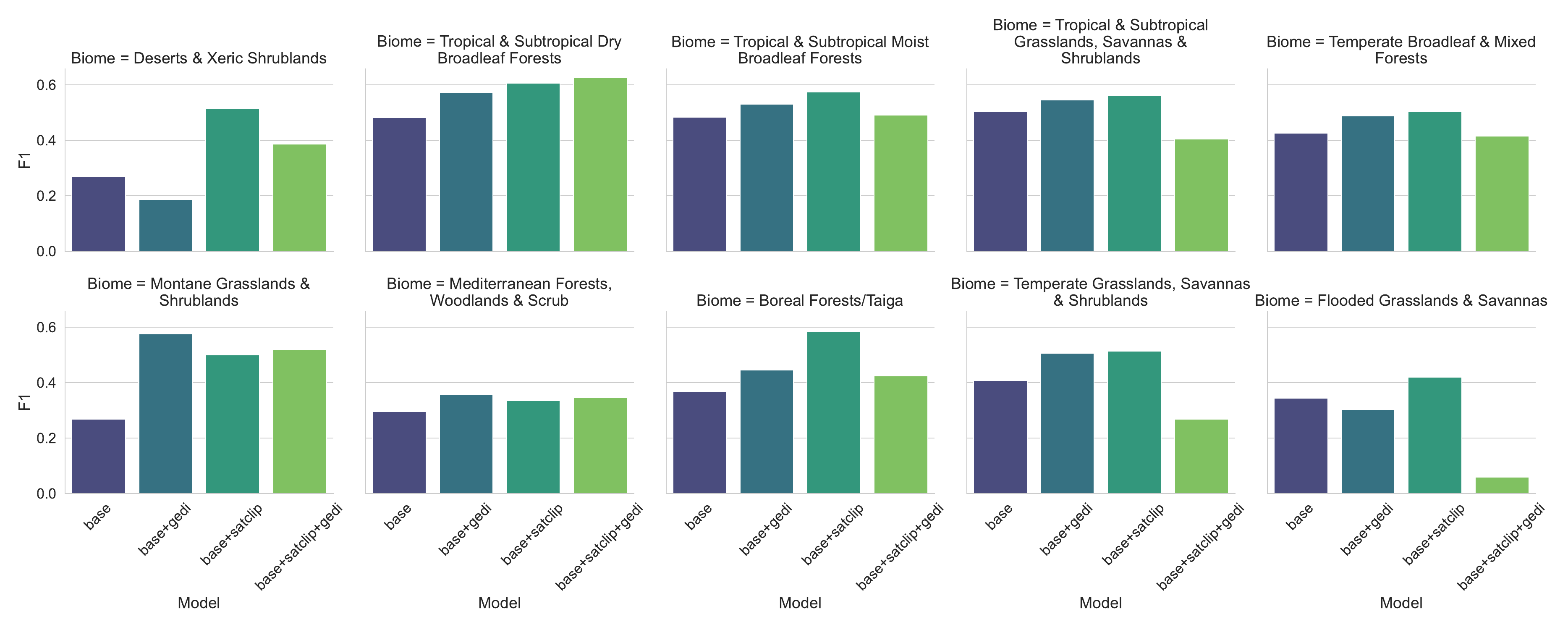}
    \caption{Performance for tree detection on different biomes, F1 scores after one-to-one matching of predicted and labeled points with a maximum distance of 15m (5 pixels).}
    \label{fig:det_quantitative}
\end{figure}

We report an acceptable average performance for mapping trees labeled directly on PlanetScope imagery. The best performing model is the version with SatCLIP embeddings (overall F1 $=0.51$, average F1 per biome $=0.51$). The SatCLIP+GEDI model performs worse (overall F1 $=0.39$, average F1 per biome $=0.39$), indicating that negative supervision does not enhance individual tree detection. However, adding GEDI supervision does improve results (overall F1 $=0.47$, average F1 per biome $=0.45$) compared to the \textit{base} model (overall F1 $=0.40$, average F1 per biome $=0.39$), although the performance variation increases across biomes. For desertic areas, GEDI supervision decreased performance significantly, which we believe is due to an over-representation of desertic areas in the GEDI set sending a negative signal not compensed by positive signals in our pseudolabels.

SatCLIP embeddings bring a clear positive effect across all biomes. With access to high-level semantics (including climatic knowledge), we hypothesize that the model can learn associations with tree density or important region-specific clues, such as visual patterns helping separate bushes from trees in grasslands.

Overall detection performance remains relatively low, compared to F1 scores usually found in tree detection studies, e.g. 0.74 for urban trees with aerial imagery \cite{ventura_individual_2024} or 0.86 for tropical forests with WorldView imagery \cite{g_braga_tree_2020}. To our knowledge, this study is the first to perform individual tree detection beyond national scale, as such it is unclear what would constitute a "good" F1 score. Nevertheless, we believe there are three main factors to take into consideration for performance evaluation. First, our model remains mechanically biased by its training dataset, and as such might not perform well in areas where visual patterns never seen during training are found. There is almost no overlap between our pseudolabel training set (Figure~\ref{fig:mainfig}) and our  evaluation set for tree detection (Figure~\ref{fig:dettestset}). Second, PlanetScope images have notorious quality issues \cite{anger_assessing_2019}, with visual artifacts producing confusing patterns, and blurred areas with a lower effective ground resolution. Third, we noticed that the distribution of heatmap values varies greatly between areas. Model confidence is linked to the volume and quality of training data, its intrinsic capacity, as well as the quality of the input data. In practice, this makes the choice of detection parameters (threshold and minimum distance between points) arduous. We partially abstract away this issue with biome-specific thresholds, but acknowledge that higher performance could surely be reached with suited parameter selection heuristics for each image.

\begin{figure}[h]
    \centering
    \includegraphics[width=\linewidth]{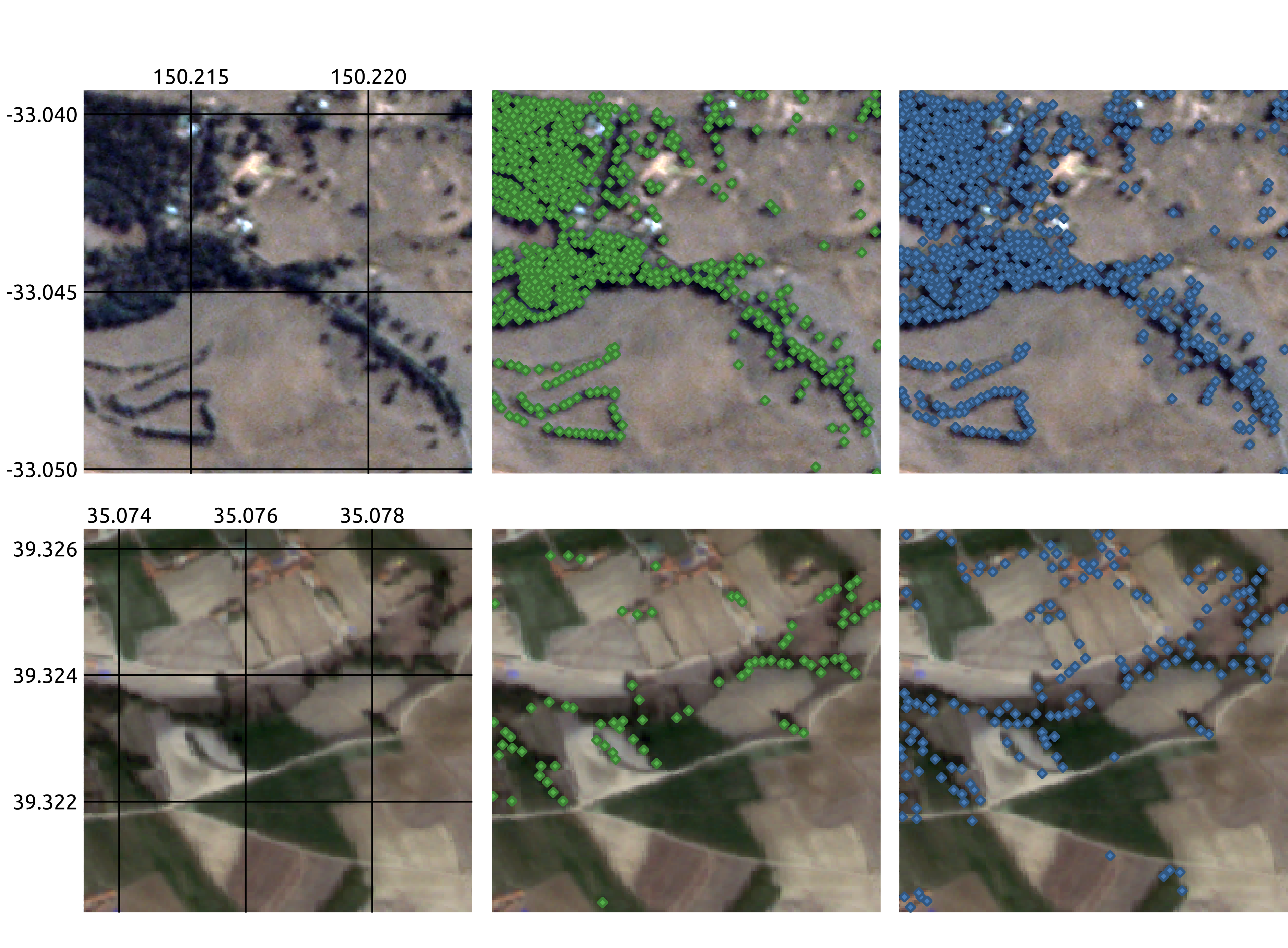}
    \caption{Qualitative examples of tree detection. First column: PlanetScope imagery (3m GSD). Second column: manual labels made through photointerpretation. Third column: model predictions, with a manually selected threshold.}
    \label{fig:det_quali}
\end{figure}

Despite these challenges, we note that our best performing model (\textit{base+satclip}) is spatially consistent, with predictions mostly situated in areas where there is vegetation, and most importantly generalizes well to unseen areas. We plot visual examples of individual detections in Figure~\ref{fig:det_quali}, on continents that were not covered by our pseudolabel set (Asia, Australia). Gaussian modelling creates clear peaks in the heatmaps, which serves as a strong clue for detecting individual trees. While we are limited by PlanetScope's 3m resolution and cannot map trees smaller than the pixel size, we believe this validates anchor-free detection with heatmap regression as a strong alternative to bounding box approaches for tree mapping, with both high performance and efficient computing making it suitable for large-scale studies.

\subsection{Cover Mapping}

\newcolumntype{x}[1]{>{\centering\let\newline\\\arraybackslash\hspace{0pt}}p{#1}}
\begin{table}[]
    \centering
    \begin{tabular}{p{0.5cm}p{2cm}|x{1.5cm}x{1.5cm}x{1.5cm}x{1.5cm}x{1.5cm}x{1.5cm}}
        & & \rotatebox{55}{WorldCover} & \rotatebox{55}{Lang et al.} & \rotatebox{55}{Tolan et al.} & \rotatebox{55}{Pauls et al.} & \rotatebox{55}{JRC} & \rotatebox{55}{Ours}\\
        \hline
        \multirow{9}{*}{\rotatebox{90}{Global threshold}} & Denmark & 0.49 & 0.61 & 0.68 & 0.75 & 0.18 & 0.76\\
         & Estonia & 0.64 & 0.74 & 0.82 & 0.90 & 0.46 & 0.89\\
         & Finland & 0.27 & -1.31 & 0.77 & 0.50 & 0.20 & 0.62\\
         & France & 0.00 & 0.04 & 0.32 & -0.26 & -0.28 & -0.16\\
         & Kenya & -2.81 & -2.98 & 0.34 & -2.19 & -1.56 & -10.67\\
         & Latvia & 0.69 & 0.12 & 0.48 & 0.32 & 0.59 & 0.60\\
         & Netherlands & -0.25 & -0.29 & 0.47 & -0.16 & 0.20 & 0.62\\
         & USA & 0.64 & 0.75 & 0.92 & 0.76 & 0.67 & 0.90\\
         & \textit{overall} & 0.77 & 0.67 & 0.78 & 0.78 & 0.66 & 0.83\\
         \hline
        \multirow{9}{*}{\rotatebox{90}{Specific threshold}} & Denmark & - & 0.61 & 0.78 & 0.84 & - & 0.76\\
         & Estonia & - & 0.74 & 0.83 & 0.90 & - & 0.90\\
         & Finland & - & -0.48 & 0.77 & 0.65 & - & 0.66\\
         & France & - & 0.21 & 0.50 & 0.40 & - & 0.68\\
         & Kenya & - & -2.16 & 0.34 & -1.92 & - & 0.50\\
         & Latvia & - & 0.72 & 0.48 & 0.75 & - & 0.74\\
         & Netherlands & - & 0.72 & 0.82 & 0.66 & - & 0.71\\
         & USA & - & 0.75 & 0.92 & 0.88 & - & 0.90\\
         & \textit{overall} & - & 0.82 & 0.80 & 0.90 & - & 0.90\\
    \end{tabular}
    \caption{Quantitative comparison of tree cover mapping against aerial lidar (3m threshold), R$^2$ scores of predicted cover fraction (higher is better). We distinguish using a fixed threshold for all lidar sources, from using a specific threshold for each source. In both cases, best thresholds are selected on a holdout set.}
    \label{tab:cover_quanti}
\end{table}

\begin{figure}[h]
    \centering
    \includegraphics[width=0.8\linewidth]{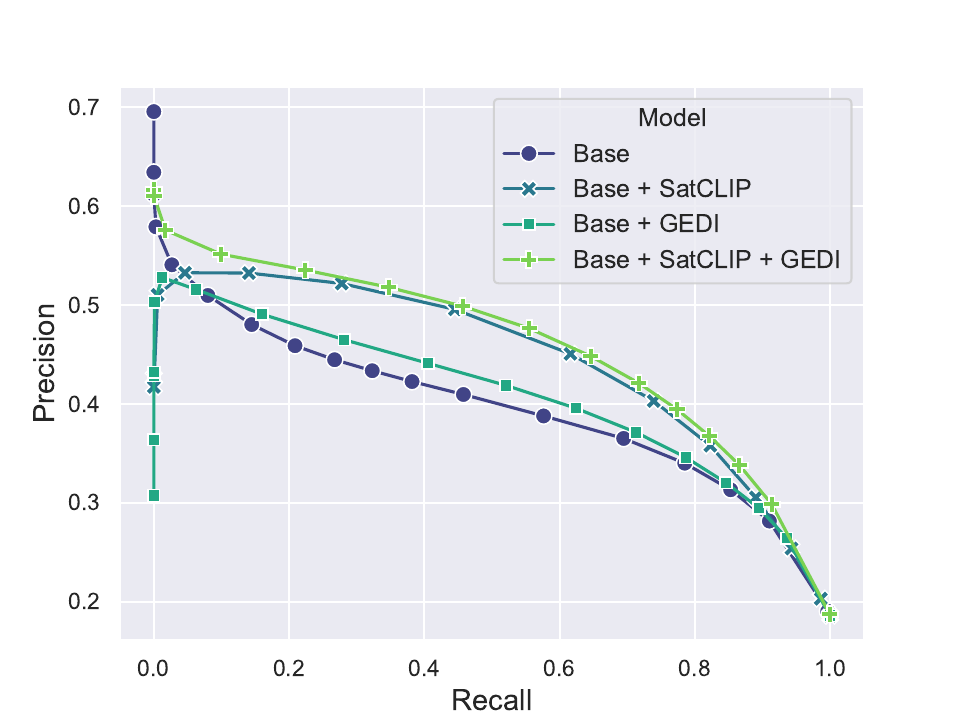}
    \caption{Precision-Recall curves for tree cover mapping against aerial lidar (3m height threshold), evaluation at pixel level.}
    \label{fig:prec_recall_cover}
\end{figure}

The \textit{base} model maps tree cover above 3m with an overall F1 score of 0.48. Adding SatCLIP embeddings significantly improves performance, with an F1 score of 0.52 and higher precision and recall at most regimes (see Figure~\ref{fig:prec_recall_cover}). Negative GEDI supervision has a small but consistent positive impact (F1$=0.53$). 

While our models were not trained for mapping binary tree cover, Gaussian modelling shows a strong potential for cover mapping with simple thresholding. The threshold directly adjusts the precision/recall balance. As for individual tree detection, performance remains strongly influenced by threshold selection, model confidence and data quality. For pixel-level evaluation in particular, co-registration issues can reduce metrics drastically, thus our efforts to collect data across various sources and ensure that we keep the highest quality for evaluation.

\begin{figure}
    \centering
    \includegraphics[width=\linewidth]{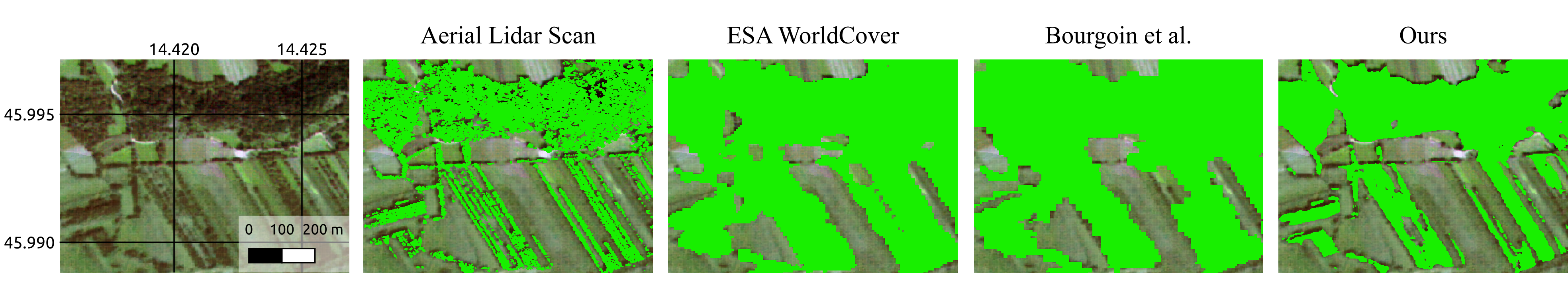}
    \caption{Qualitative comparison with cover mapping products against canopy cover from thresholded aerial lidar height scans. Left to right: PlanetScope imagery, 50cm GSD canopy height model binarized with 3m threshold, WorldCover, JRC, ours (\textit{base+satclip}, 3m GSD).}
    \label{fig:cover_quali_binary}
\end{figure}

We plot visual examples for cover mapping against ESA's WorldCover and JRC's Global Forest Cover in Figure~\ref{fig:cover_quali_binary}. The differences in resolution is clear, with 10m products producing large positive blobs, while our predictions capture finer details. 

Our predictions compare favorably to existing products, on most CHM sources (Table~\ref{tab:cover_quanti}). We report a higher R$^2$ score when using a universal threshold, and a R$^2$ score matching the best performing product (Pauls et al.) when using specific thresholds. We note that height mapping products are sensitive to height thresholds, similarly to our models with heatmap confidence thresholds. This highlights the challenge of balancing model confidence across areas for large-scale mapping, which is not solved by using global and homogenous GEDI supervision.

\begin{figure}
    \centering
    \includegraphics[width=\linewidth]{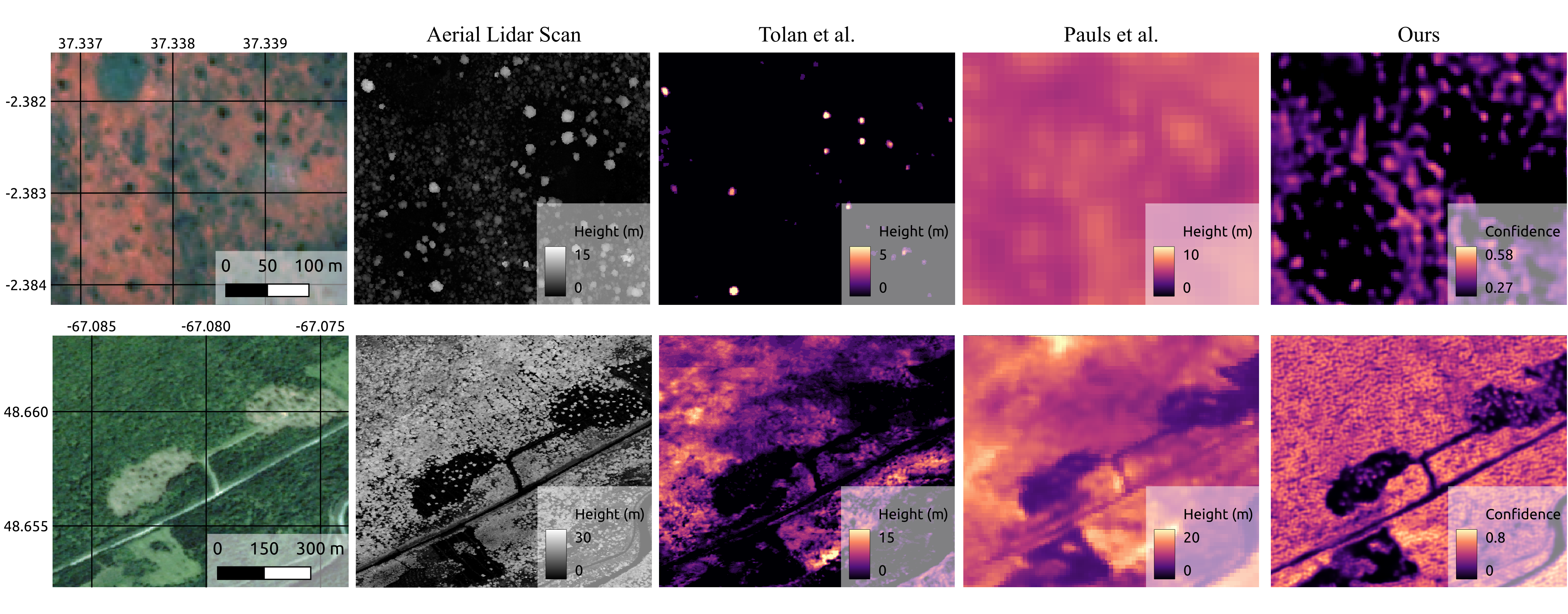}
    \caption{Qualitative comparison with height mapping products against canopy height from aerial lidar. Left to right: PlanetScope imagery, 50cm GSD canopy height model, Tolan et al. height (1m GSD), Pauls et al. height (10m GSD), ours (\textit{base+satclip}, 3m GSD).}
    \label{fig:cover_quali}
\end{figure}

On visual inspection (Figure~\ref{fig:cover_quali}), Tolan et al. maps individual trees finely with sharp activations, but tends to underpredict. In comparison, Sentinel-based products such as Pauls et al. produce well-calibrated height maps that translate to accurate aggregated cover with suited threshold selection, but lacks the spatial resolution needed to identify isolated trees and detailed patterns. Our approach offers a viable alternative, at higher resolution and maintaining correct performance. Learning centerness with a Gaussian kernel for each tree leads to heatmaps that tend to activate equally regardless of tree density, which allows mapping trees both inside and outside forests.

\subsection{Spatial Uncertainty}

\begin{figure}
    \centering
    \includegraphics[width=0.8\linewidth]{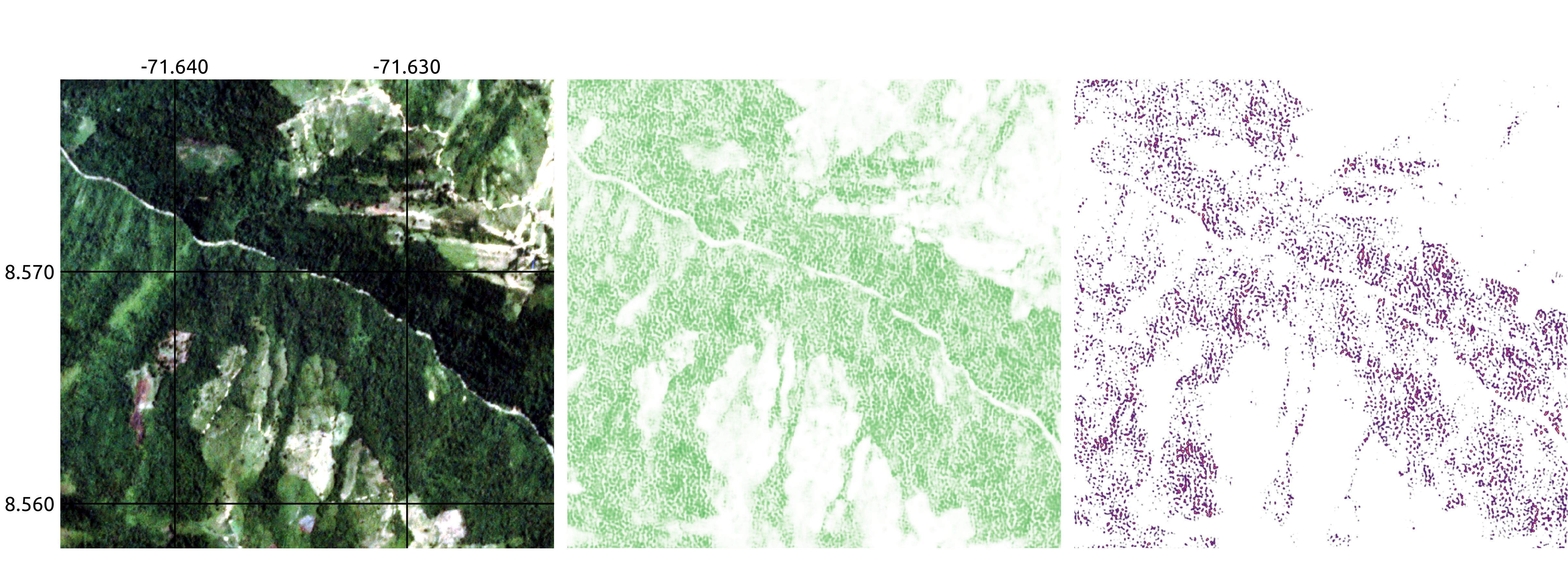}
    \caption{Heatmap (second column) and spatial uncertainty (third column) outputs. }
    \label{fig:spatial_uncertainty}
\end{figure}

Our model's outputs consists of a heatmap, indicating the confidence that a given pixel contains a tree center, and a spatial uncertainty map, indicating the model's own assessment of how uncertain tree positions are, for each pixel. On a selected area in Figure~\ref{fig:spatial_uncertainty}, we note that spatial uncertainty activates mostly on shadow areas, where there are not enough visual clues to confidently locate tree centers. On the contrary, where there is more contrast and trees become visible, e.g. on the top right corner, spatial uncertainty is close to zero and the model outputs sharper Gaussian kernels.

\subsection{Fine-tuning}

\begin{figure}[h]
\begin{subfigure}[T]{0.45\textwidth}
    \centering
    \includegraphics[height=3cm]{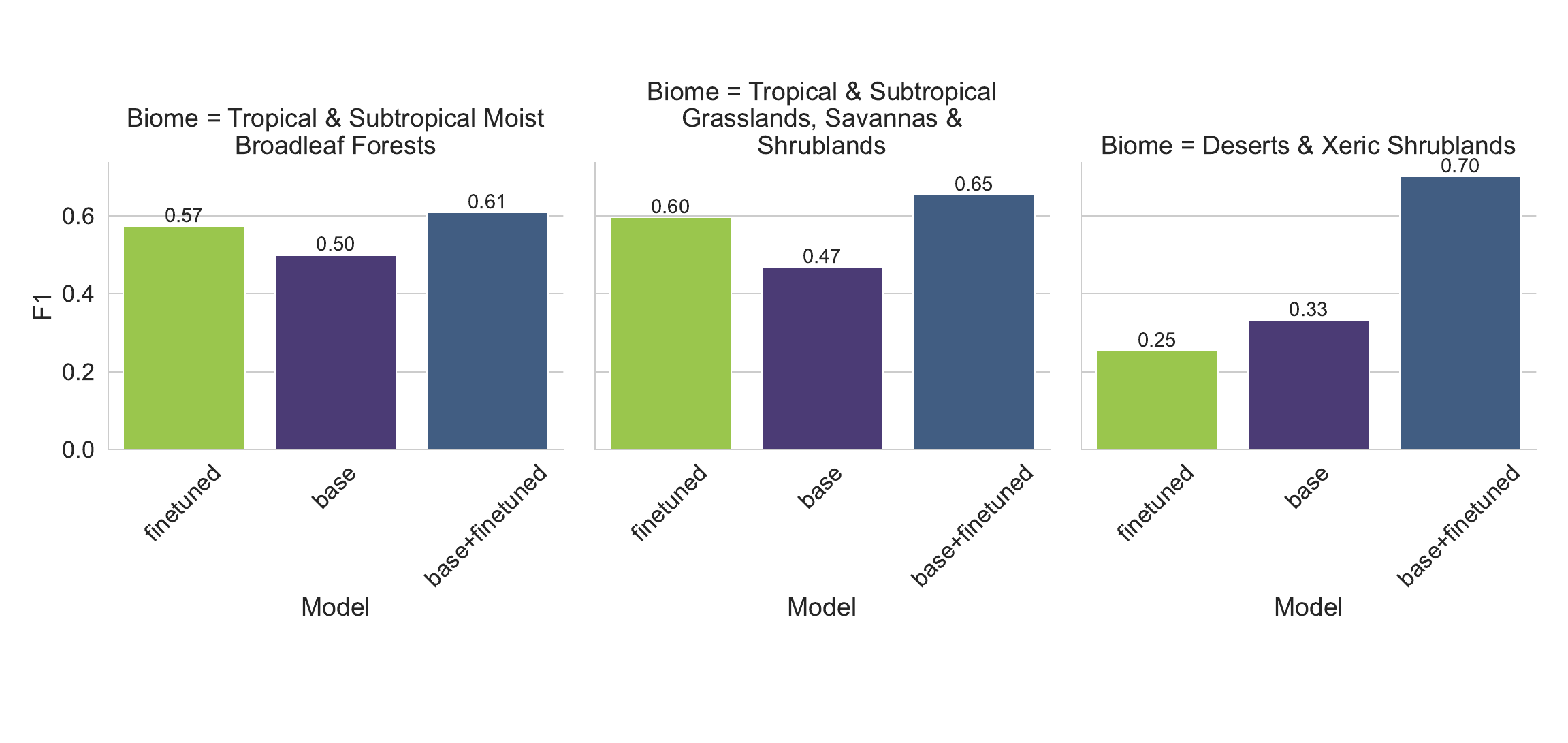}
    \caption{Quantitative evaluation of models for tree detection depending on training scheme (\textit{finetuned}: training on manual labels with ImageNet weights, \textit{base}: training with our pseudolabel dataset, \textit{base+finetuned}: finetuning the base model with manual labels).}
        \label{fig:finetuning_quanti}

\end{subfigure}
\hspace{0.05\textwidth}
\begin{subfigure}[T]{0.45\textwidth}
    \includegraphics[height=3cm]{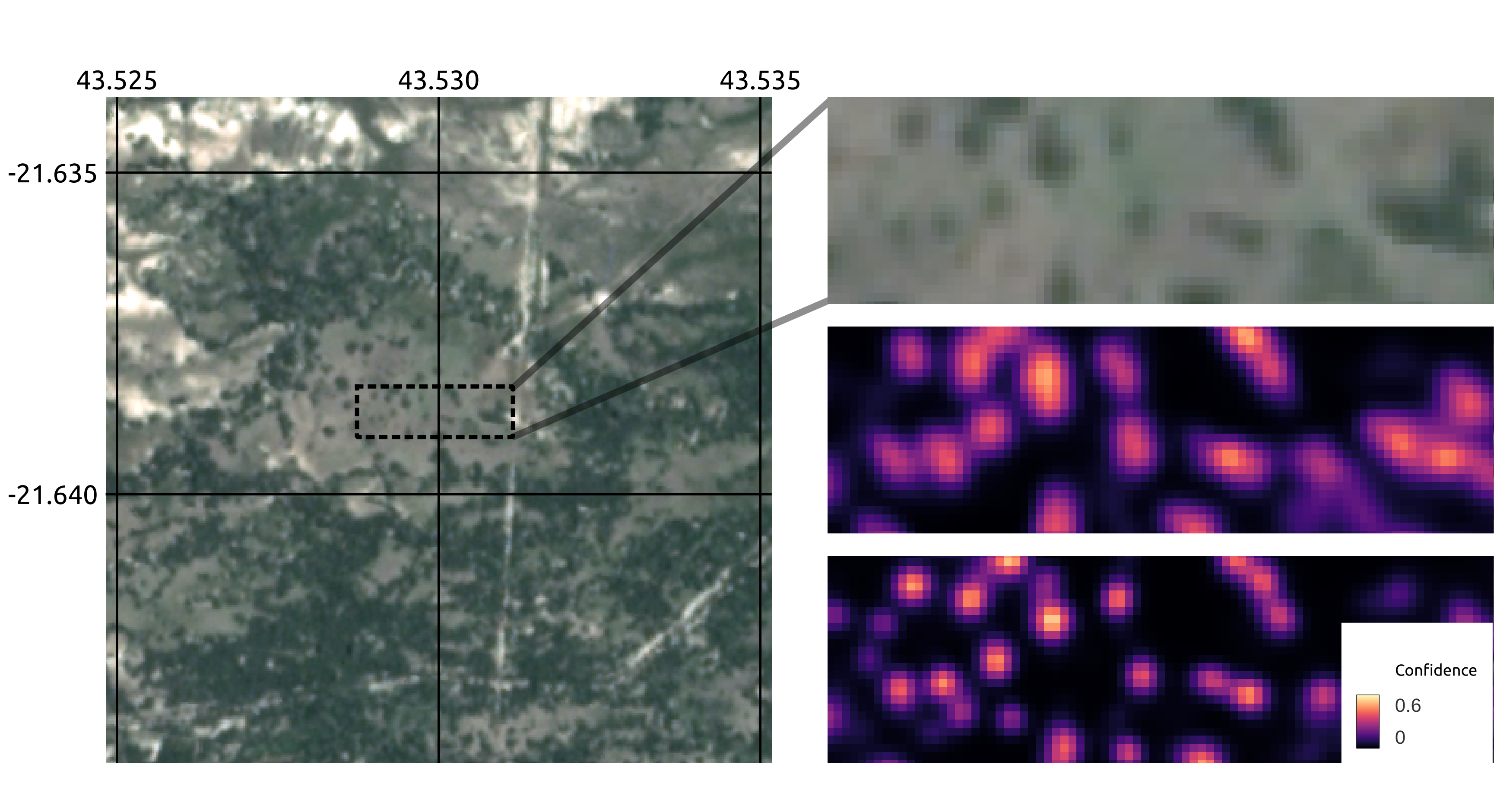}
    \caption{Predictions of the \textit{base} model (middle row) versus after finetuning (bottom row).}
    \label{fig:finetuning_quali}
\end{subfigure}
\caption{Effects of finetuning our models with manual labels.}
    \label{fig:finetuning}

\end{figure}

A common use-case for foundation models is fine-tuning them to enhance performance or adapt them to downstream tasks \cite{jakubik_foundation_2023, astruc_omnisat_2025, nedungadi_mmearth_2024}. Our models can be used similarly; our approach playing the role of a pretraining scheme that will both speed up training time and enhance performance when re-training with new labels.

We finetuned the \textit{base} model on a subset of the manual labels presented in Section~\ref{subsubsec:tree_detection}, and re-evaluated on the remaining labels. We compare this fine-tuned model to the \textit{base} model, and to a model directly trained from ImageNet weights.

The quantitative results in Figure~\ref{fig:finetuning_quanti} show a clear positive effect of pretraining on our pseudolabel dataset before fine-tuning with manual labels, compared to directly training on the manual labels. There is a significant performance margin across all biomes. Qualitatively, the heatmaps exhibit sharper Gaussian kernels (Figure~\ref{fig:finetuning_quali}), less false negatives, and a overall higher confidence towards tree centers. This underlines the complementarity between pseudolabel pretraining, which prepares the model with strong priors about tree presence regardless of the downstream task, and manual labels, which specializes the model towards tree detection, if necessary on targeted areas.

\subsection{Hyperparameter selection}

\begin{figure}[h]
\begin{subfigure}[T]{0.45\textwidth}
    \centering
    \includegraphics[height=3.5cm]{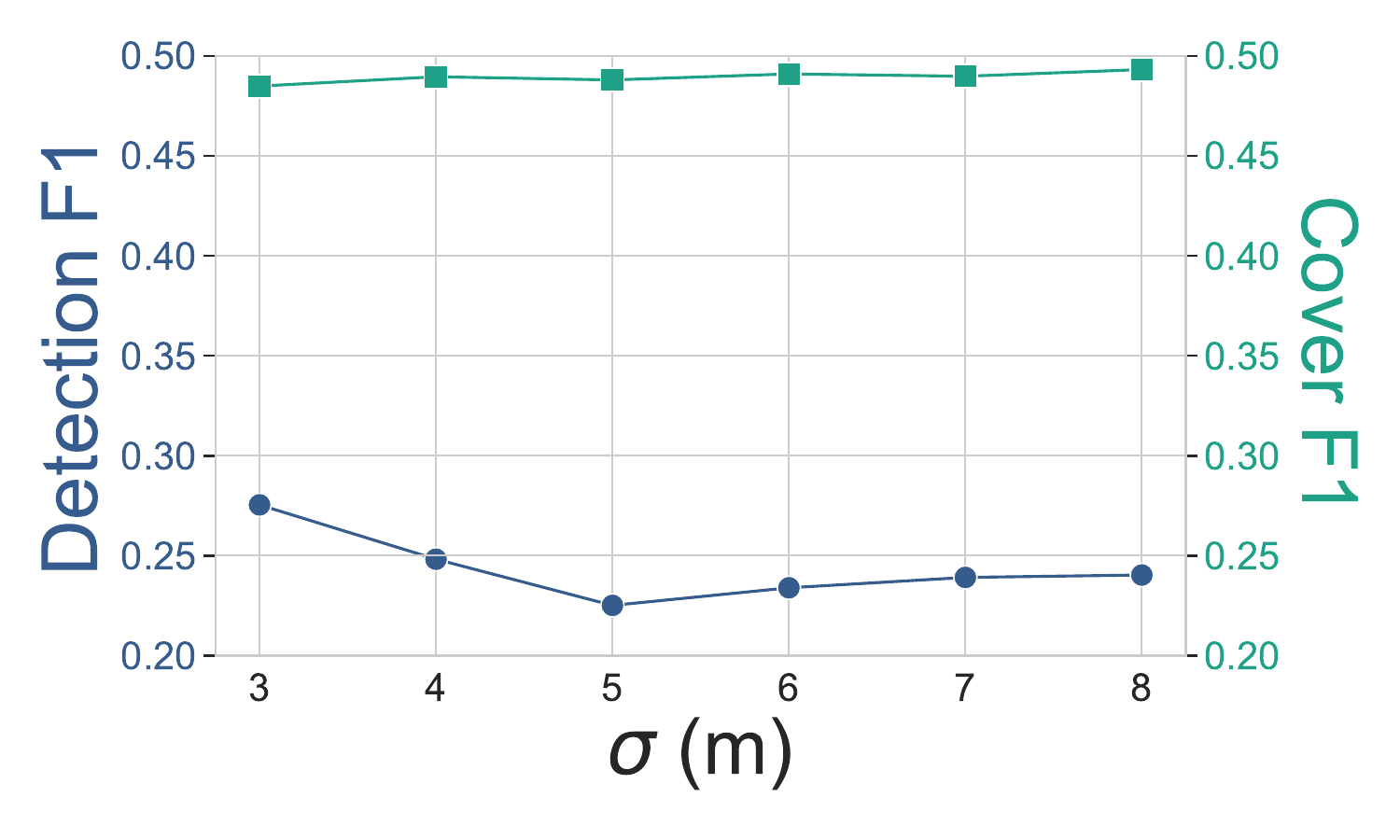}
    \caption{\label{fig:ablation_sigma}}

\end{subfigure}
\hspace{0.05\textwidth}
\begin{subfigure}[T]{0.45\textwidth}
    \includegraphics[height=3.5cm]{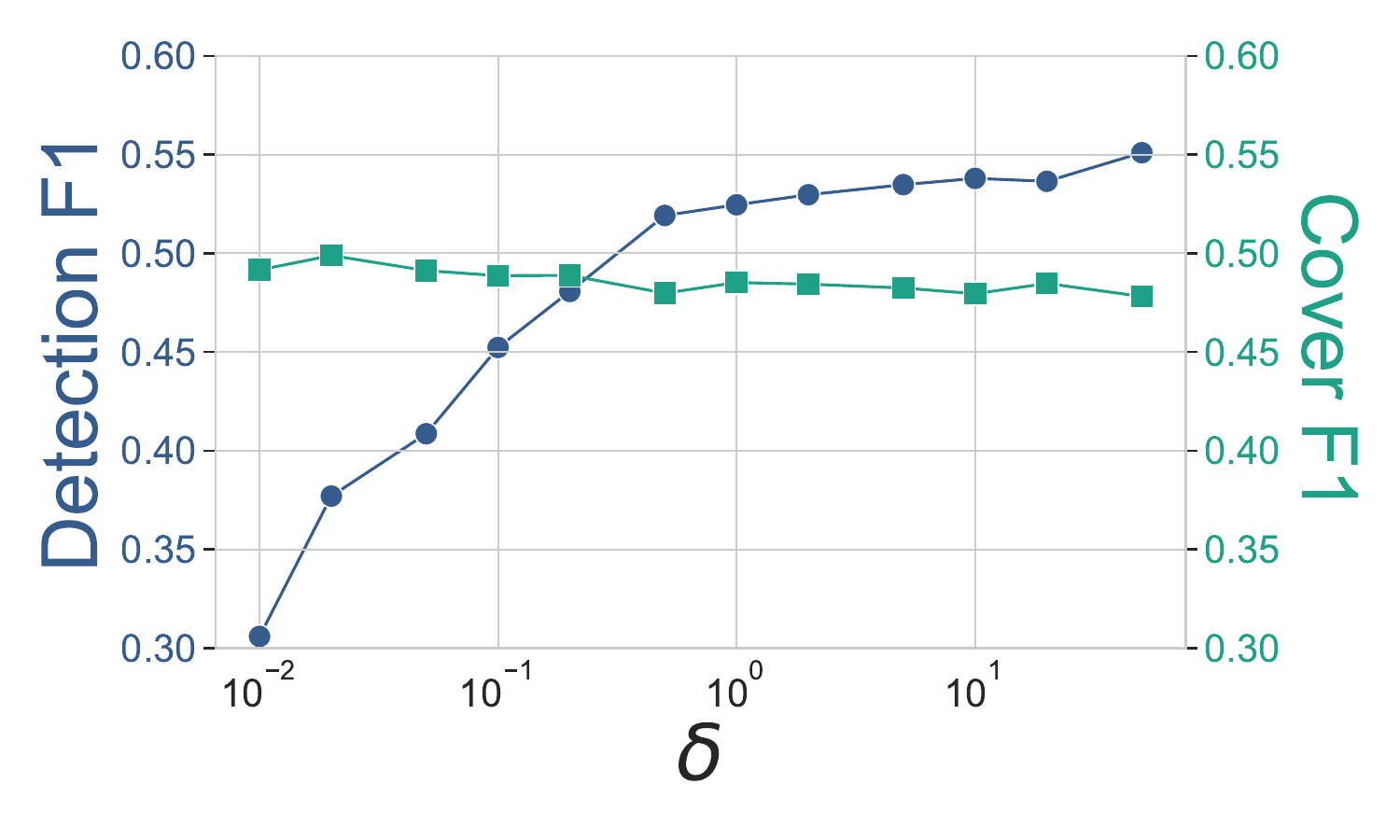}
    \caption{\label{fig:ablation_delta}}

\end{subfigure}
\caption{Impacts of hyperparameters on detection and cover mapping performance. \subref{fig:ablation_sigma} $\sigma$, minimal Gaussian size. \subref{fig:ablation_delta}$\delta$, Gaussian resizing regularization.}
    \label{fig:ablation}

\end{figure}

Our approach involves two key hyperparameters, $\sigma$, which defines the minimal Gaussian kernel size, and $\delta$ which regularizes Gaussian resizing. We report in Figure~\ref{fig:ablation} the evolution of detection and cover mapping metrics depending on these parameters, training on a subset of our pseudolabel set with the \textit{base} model. 

$\sigma$ has low influence on cover mapping performance: regardless of how wide Gaussian kernels are, it remains possible to select a threshold that will balance false positives and negatives and maintain constant performance. On the contrary, detection is more accurate with low values of $\sigma$. Higher values make Gaussian kernels "bleed" into neighbors, making peak identification less consistent. We also note that models trained with low $\sigma$ values need more time to converge, since the positive supervision for training heatmap regression becomes scarcer.

$\delta$ correlates positively with detection performance but negatively with cover mapping performance. At low values, the model can freely resize Gaussian kernels, letting to heatmaps better covering crown areas but probably ignoring smaller trees that are close to larger trees. Inversely, at high delta, each tree has the same Gaussian kernel, which emphasizes detection regardless of size but does not necessarily translate to accurate cover mapping since crown area varies greatly. 

In our experiments, we set $\sigma = 4.0$m and $\delta = 0.2$ to aim for a balance between detection and cover mapping, with reasonable training times.

\subsection{Limitations}

We acknowledge important limitations with our approach. Our framework is limited by pixel size, which means in practice that the model cannot differentiate a single large tree that covers a few pixels from a group of smaller trees covering the same area. The pseudolabels were however created with only sub-meter resolution data, and could therefore be used seamlessly with the same approach and higher resolution imagery, e.g. WorldView.

Our approach is also limited by the availability of aerial lidar, which currently is provided publicly only in few countries, mostly in Europe and North America. Although our models exhibit a high generalization capacity, if performance in a specific area is the goal, labels for that area are probably needed. Our approach was designed with large-scale mapping in mind, which ultimately means compromising performance for outlier cases (\textit{e.g.} rare biomes) to favor a balanced and high performance for the most common situations. It is however interesting to note that certain biomes for which we have barely no training data, \textit{e.g.} tropical dry broadleaf forests, still perform well, which hints that there is an interest in going large-scale that goes beyond building global maps. Efforts in data collection and curation at global scale could produce both better models overall, and better area-specific models, as is the goal of foundation models.

We noted a high variation in heatmap distribution values depending on the area, due to changes in model confidence, image quality, and visual patterns. This makes threshold selection for detection or cover mapping arduous, as different thresholds are probably needed for different areas. Optimal thresholds can be selected upon visual inspection, or through automated parameter sweeping when labeled data is available.

\section{Conclusion}

We introduced an approach to train models for individual tree mapping at large scale, using automated label mining with limited manual intervention. We showed that models trained with this framework perform competitively with existing maps for cover mapping, and can detect trees to an acceptable and homogenous level of performance, although direct comparison was not possible as there are currently no established baselines at this scale. 

We believe Gaussian modelling is a promising approach that can learn from noisy labels at scale, and is not limited to PlanetScope imagery. Spatial uncertainty modelling alleviates label noise, and gives Gaussian kernels that effectively scale with crown size. The resulting heatmaps activate where trees are found, similarly to the ubiquitous NDVI features, but with more discriminative power to separate trees from background and other vegetation. With fully convolutional UNets, predictions can quickly be obtained and merged over years for large areas. We also show that our pretrained models are strong candidates for fine-tuning, which opens the way for area-specialized models with specific labeling.

Our framework constitutes a first step towards explicit, global-scale individual tree detection, a paradigm shift in comparison with canopy cover and height mapping products. The global mapping project based on PlanetScope is ongoing at the time of the paper submission and experimental results can be viewed online \footnote{https://rs-cph.projects.earthengine.app/view/treecover}.

\printbibliography

\end{document}